\documentclass[runningheads]{llncs}
\usepackage[T1]{fontenc}
\usepackage{bbding}
\usepackage[misc]{ifsym}
\usepackage[ruled,linesnumbered]{algorithm2e}
\usepackage[colorlinks,hyperindex,breaklinks]{hyperref}
\usepackage[table]{xcolor}
\usepackage{multirow}
\usepackage{mathrsfs}
\usepackage{booktabs}
\usepackage{threeparttable}
\usepackage{amsmath}
\usepackage{graphicx}
\usepackage{bbm}
\usepackage{bbding}
\usepackage{ulem}

\begin{document}
\title{SALI: Short-term Alignment and Long-term Interaction Network for Colonoscopy Video Polyp Segmentation}
%
\author{Qiang Hu$^{\dag}$\inst{1} 
	\and
	Zhenyu Yi$^{\dag}$\inst{2} 
	\and
	Ying Zhou\inst{1}
	\and
	Fang Peng\inst{1} 
	\and
        Mei Liu\inst{3}
        \and
        Qiang Li\inst{1}\textsuperscript{\textrm{\Letter}}
        \and
	Zhiwei Wang\inst{1}\textsuperscript{\textrm{\Letter}}}

\authorrunning{Q. Hu et al.}
%
\institute{Wuhan National Laboratory for Optoelectronics, Huazhong University of Science and Technology\\
		\email{\{huqiang77, liqiang8, zww\}@hust.edu.cn}  \and
		School of Engineering and Science, Huazhong University of Science and Technology \\
		\email{yizhenyu7479@hust.edu.cn} \and
            Huazhong University of Science and Technology Tongji Medical College 
	}

\maketitle              
\def\thefootnote{$\dag$}\footnotetext{Equal contribution; \textrm{\Letter}~corresponding author.}
\begin{abstract}
Colonoscopy videos provide richer information in polyp segmentation for rectal cancer diagnosis.
However, the endoscope's fast moving and close-up observing make the current methods suffer from large spatial incoherence and continuous low-quality frames, and thus yield limited segmentation accuracy. In this context, we focus on robust video polyp segmentation by enhancing the adjacent feature consistency and rebuilding the reliable polyp representation. To achieve this goal, we in this paper propose SALI network, a hybrid of Short-term Alignment Module (SAM) and Long-term Interaction Module (LIM).
The SAM learns spatial-aligned features of adjacent frames via deformable convolution and further harmonizes them to capture more stable short-term polyp representation. In case of low-quality frames, the LIM stores the historical polyp representations as a long-term memory bank, and explores the retrospective relations to interactively rebuild more reliable polyp features for the current segmentation. Combing SAM and LIM, the SALI network of video segmentation shows a great robustness to the spatial variations and low-visual cues.
Benchmark on the large-scale SUN-SEG verifies the superiority of SALI over the current state-of-the-arts by improving Dice by 2.1\%, 2.5\%, 4.1\% and 1.9\%, for the four test sub-sets, respectively.
Codes are at \href{https://github.com/Scatteredrain/SALI}{https://github.com/Scatteredrain/SALI}.

\keywords{Video Polyp segmentation  \and Long- and Short-term Modeling \and Colonoscopy.}
\end{abstract}
%
%
%
\section{Introduction}
Colorectal cancer (CRC) is a serious threat to human health worldwide. Colono-
scopy is the gold standard for detecting and treating CRC, during which a tiny endoscope provides video imaging of the patient's digestive tract for endoscopists to inspect polyps.
Since the endoscopists' skills and experience vary greatly, missing and misdiagnosis happens from time to time.
Therefore, automatic polyp segmentation on colonscopy videos is of great importance for providing timely prompts and richer information for diagnosis.

\begin{figure}[t]
\centering
\includegraphics[width=\textwidth]{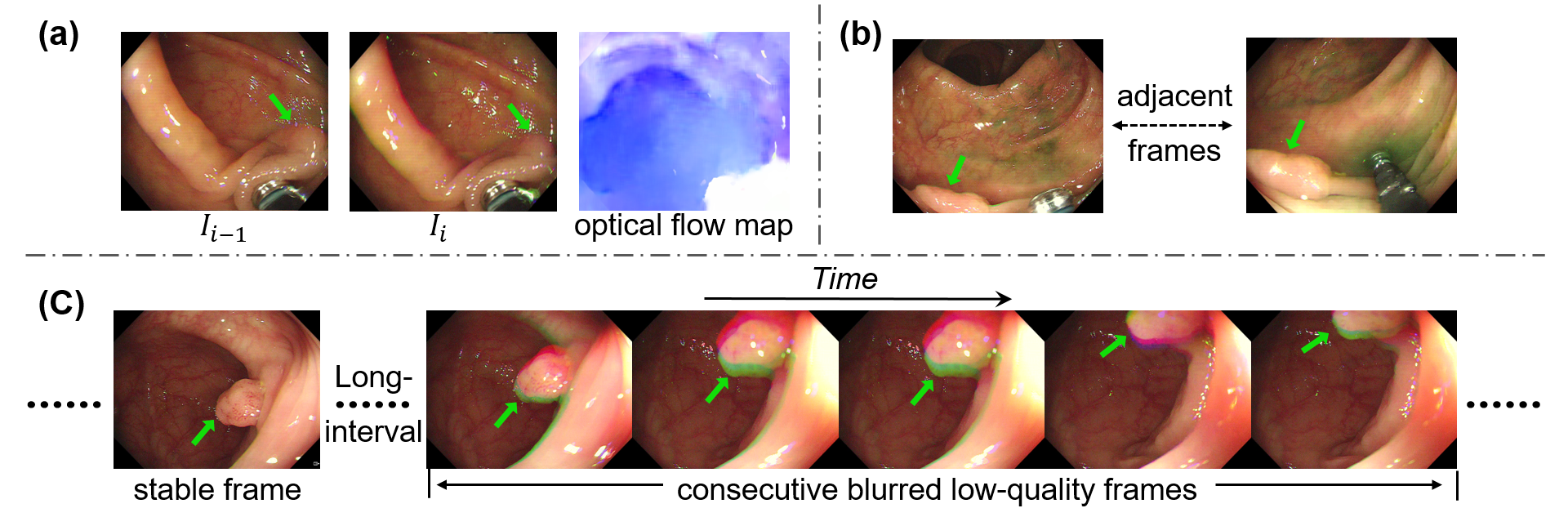}
\caption{Three challenges in polyp video segmentation. (a) The optical flow map (predicted by RAFT~\cite{teed2020raft}) can not show any object motion information. (b) Significant variations between two adjacent frames. (c) The long sequence of consecutive low-quality frames. Green arrows point to the polyps.}
\label{fig1}
\end{figure}
In the past few years, many polyp segmentation methods~\cite{fan2020pranet,wei2021shallow,dong2021polyp,zhou2023cross} for static images have been reported, but they ignore the valuable information in the time dimension and show limited performance on videos.
Fully exploiting the temporal information to model accurate feature representations of each frame is the key to video segmentation, but, this is very challenging due to the unique imaging patterns of colonoscopy.
\textbf{First}, unlike most natural scene where the camera is fixed and the object is moving, in colonoscopy videos, the camera is moving, while the object (polyp) and the background (normal tissues) are fixed instead, which results in indistinctive optical flow patterns between the polyp and normal tissues (see Fig.~\ref{fig1}(a)), and thus invalidates the optical flow-based video segmentation methods~\cite{zhang2021deep,pei2022hierarchical,yuan2023isomer}.
\textbf{Second}, the close-up observation along with fast camera trajectory causes significant frame changes in a very short time or even between two adjacent frames (see Fig.~\ref{fig1}(b)).
Therefore, existing video polyp segmentation methods~\cite{puyal2020endoscopic,ji2021progressively,ji2022video,chen2024mast}, relying on global attention blocks to directly aggregate features, would suffer from the resulting unstable short-term features.
\textbf{Third}, the complex lighting environment causes plentiful low-quality clips (see Fig.~\ref{fig1}(c)), which require long-range temporal modeling to capture reliable frames but the existing methods only consider a narrow time span~\cite{ji2021progressively,chen2024mast,puyal2020endoscopic,ji2022video} and could fail to segment at the moments full of low-quality frames.


To address the above challenges, in this paper, we propose \textbf{S}hort-term \textbf{A}lign-
ment and \textbf{L}ong-term \textbf{I}nteraction Network (SALI),
which leverages both long- and short-term receptive fields to model temporal coherence.
In a short time span, we design the Short-term Alignment Module (SAM), which first aligns features of adjacent frames by deformable convolutions to mitigate the spatial variations, and then explores the relevance of the aligned features to aggregatively construct stable short-term features.
In a large time span, we propose the Long-term Interaction Module (LIM), where a memory bank is created for remembering the historical frames and predictions.
We build the spatial-temporal correlation between the memory bank and the short-term query by a novel masked-attention block, to interactively reconstruct more reliable polyp representation even facing the quality-induced weak visual cues.

In summary, our major contributions are as follows:
\begin{itemize}
 	\item[$\bullet$] We propose a novel video polyp segmentation method SALI, which addresses the limitations of existing methods on adjacent frames of large variation and long sequences of consecutive low-quality frames. 
	\item[$\bullet$] We propose two novel modules called Short-term Alignment Module (SAM) and Long-term Interaction Module (LIM), to enhance the stability and reliability of spatial-temporal features, respectively.
	\item[$\bullet$] Benchmark results on the large-scale public dataset, i.e., SUN-SEG, demonstrate that SALI achieves a superior performance compared to other state-of-the-arts by improving Dice by 2.1\%, 2.5\%, 4.1\%, 1.9\% on the four test sub-sets, respectively.
\end{itemize}

\section{Method}
\begin{figure}[t]
\includegraphics[width=\textwidth]{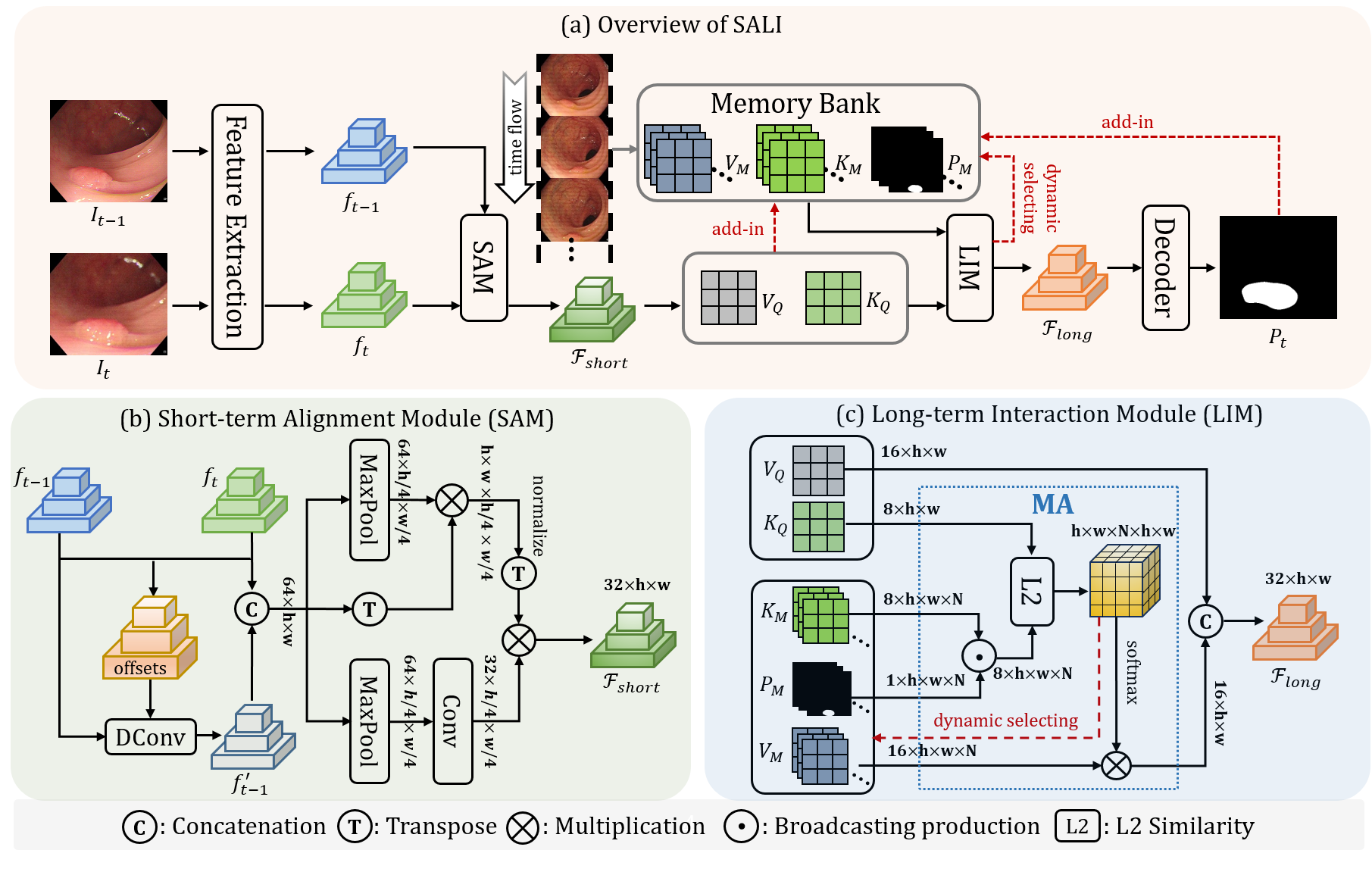}
\caption{Overview of our proposed SALI. (a) SALI proposes two modules, called short-term alignment module (SAM) and long-term interaction module (LIM), to obtain the stable and reliable spatia-temporal features. (b) SAM first aligns the adjacent features, and then constructs stable short-term features by exploring relevance. (c) LIM utilize masked-attention (MA) block to interact the short-term feature with the long-term visual cues in the memory bank to obtain reliable long-term feature.}
\label{fig2}
\end{figure}

Fig.~\ref{fig2}(a) shows the overall of SALI, and its two major modules, i.e., short-term alignment module (SAM) and long-term interaction module (LIM). In the following, we detail each module and give implementation details.

\subsection{Stable Feature Learning via Short-term Alignment Module}
SALI is an online video segmentation network, which means that it `sees' \textbf{no} future frames.
Given the current frame at the $t$ time point $I_{t} \in \mathbbm{R}^{3 \times H \times W}$, SALI first uses SAM to construct a short-term representation $\mathcal{F}_{short}$, conditioning on the previous frame $I_{t-1}$. 
Specifically, SAM employs a transformer-based feature extractor PVTv2~\cite{wang2022pvt} plus receptive field blocks (RFB)~\cite{wu2019cascaded} to extract image features for the two adjacent frames.
For each frame, three feature maps are obtained with the down-scale factor of $8$, $16$, $32$, respectively, and totally we can have $\{f_{t}^l, f_{t-1}^l\} \in \mathbbm{R}^{32 \times H/2^{l+2} \times W/2^{l+2}}$, $l=1,2,3$. 
The feature maps are processed independently at each scale $l$, and no cross-scale operation is involved until the final decoding.
Therefore, we omit the scale notation in the following.

To build short-term feature, we need to aggregate $\{f_{t}, f_{t-1}\}\in \mathbbm{R}^{32 \times h \times w}$, where $h=\frac{H}{2^{l+2}}$ and $w= \frac{W}{2^{l+2}}$ if the feature maps are at the $l$-th scale.
However, large spatial variations between them occur due to fast camera move and close-up observation, which reduces the aggregation stability.
In view of this, SAM first aligns $f_{t-1}$ with $f_{t}$ before aggregation.
The naive way is to perform registration on frames, while the deformation field is prohibitively difficult to estimation since the field of view is often non-overlapped between the two frames.
Therefore, SAM directly leans aligned features instead of explicitly estimating deformation fields.

To this end, SAM employs the deformable convolution layer (DConv)~\cite{dai2017deformable}.
Specifically, $\{f_{t}, f_{t-1}\}$ are first concatenated and fed into a $1\times1$ convolution layer $\mathcal{N}_0$ to estimate kernel offsets, which reflecting the discrepancy of the respective fields of the two frames.
The offsets are then utilized to create a $3\times3$ DConv for converting $f_{t-1}$ to $f_{t-1}'$, which can be formulated as follows:
\begin{equation}
\begin{aligned}
f_{t-1}' = \mathtt{Dconv}\left(f_{t-1},\mathcal{N}_0(\mathtt{concat}[f_{t},f_{t-1}])\right) \in \mathbbm{R}^{32 \times h \times w},
\end{aligned}
\label{eq1}
\end{equation}

$f_{t-1}'$ is expected to be gradually aligned with $f_{t}$ during the end-to-end optimization.
Therefore, we can aggregate them via concatenation and further harmonize them using a self-attention $\mathtt{Attention}$ operation plus a $3\times3$ convolution layer $\mathcal{N}_1$, which is visualized in Fig.~\ref{fig2}(b).
Finally, the short-term feature $\mathcal{F}_{short}$ can be calculated as follows:
\begin{equation}
	\mathcal{F}_{short} = \mathtt{Attention}(Q,K,V) \in \mathbbm{R}^{32 \times h \times w},
	\label{eq3}
\end{equation}
where $Q=\mathtt{concat}[f_{t},f_{t-1}']$, $K=\mathtt{maxpool}(Q)$, $V=\mathcal{N}_1(\mathtt{maxpool}(Q))$, and the window size in the max-pooling is $4\times4$.

\subsection{Reliable Feature Learning via Long-term Interaction Module} 
Technically, $\mathcal{F}_{short}$ can be utilized to decode the segmentation mask, while $\{I_{t}, I_{t-1}\}$ may contain low image quality due to the complex lighting environment, making the decoding unreliable.
To rebuild more reliable representation, SALI utilizes LIM to gain the ability of perceiving a long-term memory of polyp representations.

Specifically, LIM first utilizes a $3\times3$ convolution layer to convert the short-term feature $\mathcal{F}_{short}$ into a key-value pair, denoted as $\{K\in \mathbbm{R}^{8 \times h\times w}, V\in \mathbbm{R}^{16 \times h\times w}\}$.
As the video stream goes, the key-value pairs and the final mask predictions are continually added into a $N$-length memory bank, yielding the historical key-value set, i.e., $\{K_M, V_M\}$, and mask set, i.e, $P_M \in \mathbbm{R}^{1 \times h\times w \times N}$.
To be clearer, as shown in Fig.~\ref{fig2}(c), the key-value pair derived from the current short-term feature is denoted as $\{K_Q, V_Q\}$.

Our goal is to rebuild $\{K_Q, V_Q\}$ into a more reliable long-term feature $\mathcal{F}_{long}$ by use of the previously-seen discriminative polyp representations stored in the memory bank.
To this end, LIM introduces a masked-attention (MA) block, which utilizes the current key $K_Q$ to retrieve the most correlated historical frames by use of the polyp regions in the stored $K_M$.
Then, the MA block aggregates the retrieved frames' feature values according to the normalized correlations.
We call this process long-term interaction since the key insight is to interactively gather those reliable visual cues appearing in a large time span to rebuild the current reliable representation. 

At last, $\mathcal{F}_{long}$ is obtained by concatenating the current feature value with the interacted one, which is formulated as follows:
\begin{equation}
	\mathcal{F}_{long} = \mathtt{concat}\left[ V_Q, {V_M} \times \mathtt{softmax}\left(\frac{ \mathcal{S}( {K_Q,K_M \odot P_M)}} {\sqrt{d_k}} \right) \right] \in \mathbbm{R}^{32 \times h \times w},
	\label{eq4}
\end{equation}
where $d_k$ is a scale factor~\cite{vaswani2017attention}, $\mathcal{S}({\cdot})$ computes the L2 similarity~\cite{cheng2021rethinking}, and $\odot$ denotes broadcasting production.

To balance the memory cost, we set the maximum memory length $N$ to 35, and add the key-value pair into the memory bank every 5 frames.
If the memory bank is full, a dynamic selecting strategy will be evoked to make room for a new entering element.
Specifically, the least relevant historical frame indicated by the L2 similarities in Eq.~\ref{eq4} will be deleted.

\subsection{Segmentation and Training details} 
For segmentation, SALI first obtains $\{\mathcal{F}_{long}^l\}$ from all the three scales, and then leverages the partial decoder~\cite{wu2019cascaded} to aggregate $\{\mathcal{F}_{long}^l\}$ to obtain a global map, and at last uses the global map to predict the final segmented map $P_{t}$ at the current time via the reverse attention~\cite{fan2020pranet}.
Then we compute the combination of cross-entropy (CE) loss and IoU loss between the segmented map $P_{t}$ and the ground truth mask $Y_t$ to optimize our model as follows:
\begin{equation}
\mathcal{L} = \mathcal{L}_{CE}(P_{t},Y_t) + \mathcal{L}_{IoU}(P_{t},Y_t)
\label{eq8}
\end{equation}

We implement SALI using PyTorch~\cite{paszke2019pytorch} and trained using a single NVIDIA GeForce RTX 4090 GPU with 24GB memory.
The ImageNet~\cite{deng2009imagenet} pre-trained weights of PVTv2~\cite{wang2022pvt} are loaded as initialization.
We reisze the input images into $352 \times 352$, and set the batch size to 14.
We use Adam~\cite{kingma2014adam} as our optimizer with weight decay of 1e-4.
We train the model for 30 epochs and set the learning rate to 1e-4.

\section{Experiments}
\subsection{Datasets and Evaluation Metrics.}
We conduct experiments on SUN-SEG~\cite{ji2022video}, the largest polyp video segmentation dataset.
Following previous works~\cite{chen2024mast,ji2022video}, the train set contains 112 video clips with a total of 19,544 frames, and the test set is divided into four sub-test sets, i.e, SUN-SEG-Seen-Easy (33 clips/4,719 frames), SUN-SEG-Seen-Hard (17 clips/3,882 frames), SUN-SEG-Unseen-Easy (86 clips/12,351 frames), and SUN-SEG-Unseen-Hard (37 clips/8,640 frames).
Easy/Hard indicates that difficult levels to be segmented of the samples, Seen indicates that the clips are from the same video as the train set but do not overlap, and Unseen indicates that the clips are from videos that do not overlap with the train set.

For comprehensive comparison, we employ six metrics to evaluate the results, i.e., Dice, structure-measure ($S_\alpha$)~\cite{fan2017structure}, enhanced-alignment measure ($E_{\phi}^{mn}$)~\cite{fan2021cognitive}, sensitivity (Sen), F-measure ($F_\beta^{mn}$)~\cite{achanta2009frequency}, and weighted F-measure ($F_\beta^w$)~\cite{margolin2014evaluate}, respectively, similar to previous works~\cite{chen2024mast,ji2022video}.


\begin{table}[t]
\centering
\caption{Quantitative comparison on SUN-SEG dataset. The best performance is marked in bold.}
{
\begin{tabular}
{
p{0.4cm}<{\centering}
lcccccccccccc}

\toprule
    \multicolumn{2}{c}{ \multirow{2}{*}{Method} }& \multicolumn{6}{c}{SUN-SEG-Easy} & \multicolumn{6}{c}{SUN-SEG-Hard} 
    \\
     \cmidrule(lr){3-8}  \cmidrule(lr){9-14}  
    &  &$S_\alpha$ &$E_{\phi}^{mn}$ &$F_\beta^w$ 
    &$F_\beta^{mn}$ &Sen &Dice 
    &$S_\alpha$ &$E_{\phi}^{mn}$ &$F_\beta^w$ 
    &$F_\beta^{mn}$ &Sen &Dice 
    \\
    \midrule
    \multicolumn{1}{l|}{\multirow{6}{*}{\rotatebox{90}{Seen}}}
        &2/3D~\cite{puyal2020endoscopic}
        &89.5 &90.9 &81.9 &85.3 &80.8 &\multicolumn{1}{l|}{85.6} 
        &84.9 &86.8 &75.3 &80.5 &72.6 &80.9
        \\
    \multicolumn{1}{l|}{} 
        &PNS-Net~\cite{ji2021progressively}
        &90.6 &91.0 &83.6 &86.0 &82.7 &\multicolumn{1}{l|}{86.1}
        &87.0 &89.2 &78.7 &82.2 &77.4 &82.3
        \\
    \multicolumn{1}{l|}{} 
        &PNS+~\cite{ji2022video}
        &91.7 &92.4 &84.8 &87.8 &83.7 &\multicolumn{1}{l|}{88.8} 
        &88.7 &92.9 &80.6 &84.9 &78.0 &85.5
        \\
    \multicolumn{1}{l|}{} 
        &FLA-Net~\cite{lin2023shifting}
        &90.6 &92.2 &84.6 &86.7 &85.1 &\multicolumn{1}{l|}{87.5}
        &85.9 &89.2 &77.8 &81.0 &78.5 &80.9
        \\
    \multicolumn{1}{l|}{} 
        &SLT-Net~\cite{cheng2022implicit}
        &92.7 &96.1 &89.4 &91.4 &88.8 &\multicolumn{1}{l|}{90.6}
        &89.4 &94.3 &84.7 &87.4 &85.1 &86.6 
        \\
    \multicolumn{1}{l|}{} 
        &\textbf{SALI (Ours)}
        &\textbf{94.4} &\textbf{97.2} &\textbf{91.0} &\textbf{92.5} &\textbf{92.4} &\multicolumn{1}{l|}{\textbf{92.7}} 
        &\textbf{91.6} &\textbf{95.2} &\textbf{86.5} &\textbf{88.9} &\textbf{89.8} &\textbf{89.1}
        \\
    \midrule \midrule
    \multicolumn{1}{l|}{ \multirow{7}{*}{\rotatebox{90}{Unseen}}}
        
        &2/3D~\cite{puyal2020endoscopic}
        &78.6 &77.7 &65.2 &70.8 &60.3 &\multicolumn{1}{l|}{72.2} 
        &78.6 &77.5 &63.4 &68.8 &60.7 &70.6
        \\
    \multicolumn{1}{l|}{} 
        &PNS-Net~\cite{ji2021progressively}
        &76.7 &74.4 &61.6 &66.4 &57.4 &\multicolumn{1}{l|}{67.6} 
        &76.7 &75.5 &60.9 &65.6 &57.9 &67.5
        \\
    \multicolumn{1}{l|}{} 
        &PNS+~\cite{ji2022video}
        &80.6 &79.8 &67.6 &73.0 &63.0 &\multicolumn{1}{l|}{75.6} 
        &79.7 &79.3 &65.3 &70.9 &62.3 &73.7
        \\
    \multicolumn{1}{l|}{} 
        &FLA-Net~\cite{lin2023shifting}
        &72.2 &69.7 &54.7 &59.7 &50.6 &\multicolumn{1}{l|}{63.6} 
        &72.1 &70.1 &54.0 &59.2 &52.2 &62.8
        \\
   \multicolumn{1}{l|}{}  
        &SLT-Net~\cite{cheng2022implicit}
        &84.8 &89.3 &77.5 &81.7 &74.7 &\multicolumn{1}{l|}{79.2} 
        &84.4 &90.4 &75.7 &79.5 &76.0 &78.1
        \\
    \multicolumn{1}{l|}{} 
        &MAST~\cite{chen2024mast}
        &84.5 &89.8 &77.0 &81.9 &75.5 &\multicolumn{1}{l|}{78.4} 
        &86.1 &91.4 &77.7 &81.6 &81.1 &80.3
        \\
    \multicolumn{1}{l|}{} 
        &\textbf{SALI (Ours)}
        &\textbf{87.0} &\textbf{92.0} &\textbf{79.4} &\textbf{83.1} &\textbf{81.1} &\multicolumn{1}{l|}{\textbf{82.5} }
        &\textbf{87.4} &\textbf{92.0} &\textbf{79.0} &\textbf{82.2} &\textbf{83.0} &\textbf{82.2}
        \\
\bottomrule
\end{tabular}
}

\label{table1}
\end{table}

\subsection{Comparisons with State-of-the-art Methods}
We compare SALI with six video-based state-of-the-art methods~\cite{puyal2020endoscopic,ji2021progressively,ji2022video,lin2023shifting,cheng2022implicit,chen2024mast}.
To ensure fairness, we acquire the segmentation results of these methods except MAST~\cite{chen2024mast} by utilizing their publicly available implementations.
MAST has no code but provides its predictions on two unseen sub-test sets.

\begin{figure}[t]
\centering
\includegraphics[width=\textwidth]{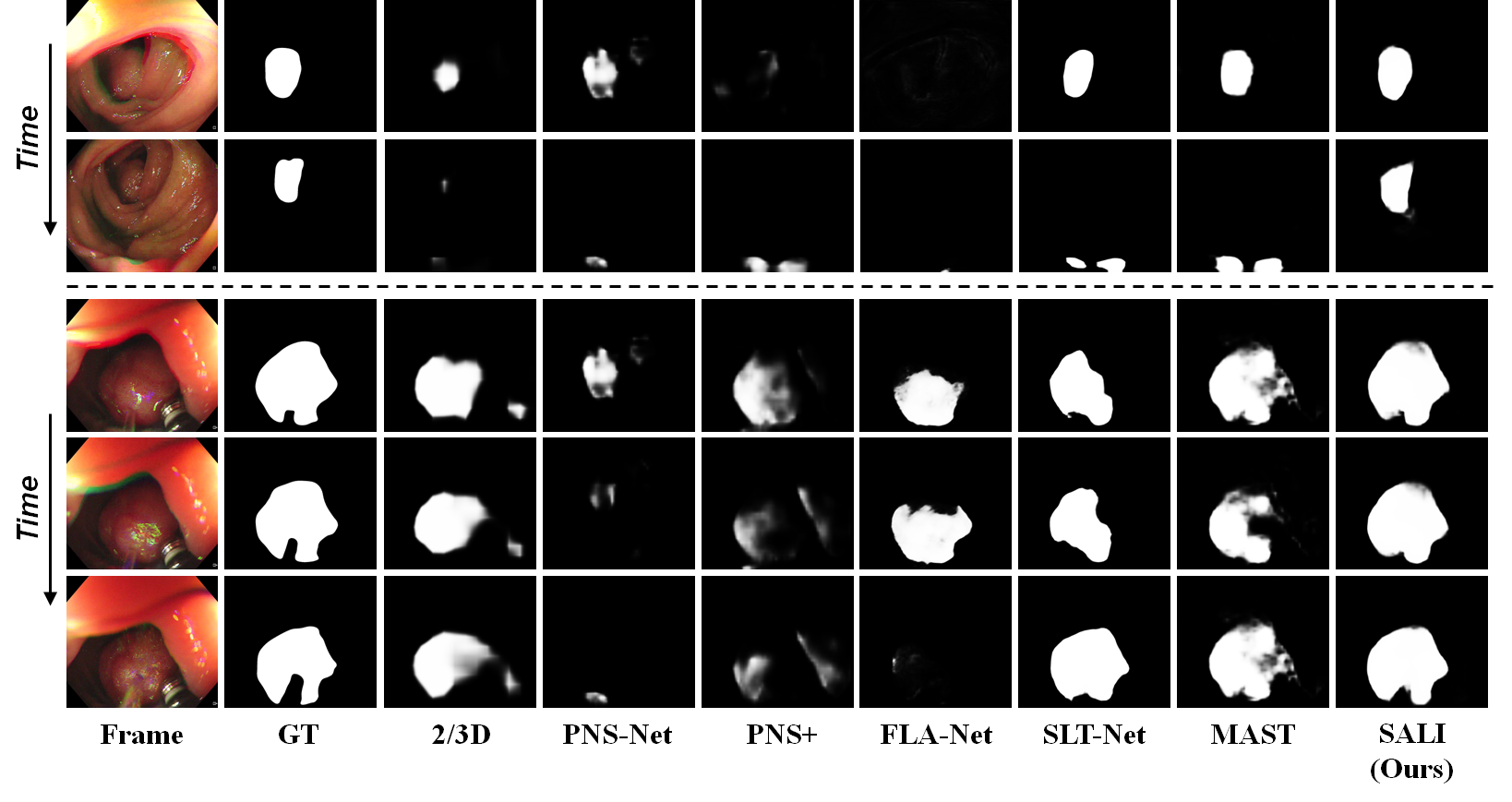}
\caption{The visualization results of different methods on two challenge cases. The upper case: significant variations between adjacent frames; the lower case: a sequence of consecutive low-quality frames.}
\label{fig3}
\end{figure}

\subsubsection{Quantitative Comparisons.}
\vspace{-0.3cm}
Table~\ref{table1} presents the quantitative comparison results.
It can be seen that our method achieves the best performance on all the metrics of the four sub-test sets.
Specifically, on the seen sets, SALI surpasses the second-best method, i.e., SLT-Net, by 2.1\% and 2.5\% in terms of Dice on SUN-SEG-Seen-Easy and SUN-SEG-Seen-Hard, respectively, which suggests that SALI has a strong learning ability to segment polyps.
On the unseen sets, our SALI exceeds the second-best method, i.e., MAST, by an increase of Dice by
4.1\% and 1.9\% on SUN-SEG-Uneen-Easy and SUN-SEG-Uneen-Hard, respectively.
This demonstrates that a better generalization capability of SALI than other methods, which is attributed to our memory-based long- and short-term interaction strategy of SALI.

\subsubsection{Qualitative Comparisons.}
\vspace{-0.3cm}
In Fig.~\ref{fig3}, we visualize the segmentation results of different methods on two challenging cases, i.e, significant variations between adjacent frames (the upper case) and a sequence of consecutive low-quality frames (the lower case).
As shown in Fig. \ref{fig3}, other methods all fail to segment polyps in these two cases.
In contrast, SALI accurately segments polyps and the results are stable in the time flow, which verify that our method can effectively segment polyps in both challenge cases.

\subsection{Ablation Study}
\begin{table}[t]
    \centering
\caption{Ablation study of two key components, i.e., SAM and LIM, and two of their sub-components, i.e., FA and MA.}
\scalebox{1}{
\begin{tabular}{lcccccccccccc}
	\toprule
	 \multirow{2}{*}{Methods} & \multicolumn{6}{c}{{SUN-SEG-Seen-Hard}} & \multicolumn{6}{c}{{SUN-SEG-Unseen-Hard}}  \\
  \cmidrule(lr){2-7} \cmidrule(lr){8-13}
    &$S_\alpha$ &$E_{\phi}^{mn}$ &$F_\beta^w$ 
    &$F_\beta^{mn}$ &Sen &Dice 
    &$S_\alpha$ &$E_{\phi}^{mn}$ &$F_\beta^w$ 
    &$F_\beta^{mn}$ &Sen &Dice \\
    \midrule
   SALI
    &\textbf{91.6} &\textbf{95.2} &\textbf{86.5} &\textbf{88.9} &\textbf{89.8} &\multicolumn{1}{l|}{\textbf{89.1}} 
     &\textbf{87.4} &\textbf{92.0} &\textbf{79.0} &\textbf{82.2} &\textbf{83.0} &\textbf{82.2}\\ 
     
    \midrule
    w/o SAM&90.5 &94.4 &85.1 &87.4 &87.6&\multicolumn{1}{l|}{87.7}
    & 85.6  &90.4 &76.6 &79.7 &78.7 &79.9\\
    w/o LIM&90.2  &94.1 &85.2 &87.6 &87.0 &\multicolumn{1}{l|}{87.4}
    &84.6  &88.8 &75.2 &78.9 &76.7 &78.5\\
    w/o SAM+LIM&88.8  &93.1 &83.1 &86.4 &83.4 &\multicolumn{1}{l|}{85.5 }
       & 81.1  &86.3 &69.8 &74.7 &67.9 &74.9 \\
    
    \midrule
    
    w/o MA 
    & 88.9  &92.3 &82.4 &84.6 & 85.1 & \multicolumn{1}{l|}{84.8}
    & 85.7  &89.4 &76.7 &80.5 & 78.2 & 80.8
    \\
     w/o FA
    & 90.6  &93.9 &85.0 &87.5 & 87.5 & \multicolumn{1}{l|}{87.6}
    & 86.2  &91.3 &77.8 &81.2 & 80.2 & 81.2
    \\
	\bottomrule
\end{tabular}
\label{table2}
}
\end{table}
To verify the effectiveness of two key components, i.e., SAM and LIM, and two of their sub-components, i.e., feature alignment (FA) and masked-attention (MA) block, we train four variants of SALI by disabling them separately and a image-based variant with only the encoder-decoder structure.
Note that, we disabling MA by directly computing the similarity between $K_Q$ and $K_M$ in Fig.~\ref{fig2}(c).
The comparison results of these method on SUN-SEG-Seen-Hard and SUN-SEG-Unseen-Hard are listed in Table~\ref{table2}.

Comparing the first four rows in Table~\ref{table2}, it can be seen that disabling SAM or LIM can both cause the performance degradation, while disabling both of them results in further degradation.
Specifically, disabling SAM degrades Dice by 1.4\% and 2.3\% on SUN-SEG-Seen-Hard and SUN-SEG-Unseen-Hard, respectively, disabling LIM degrades Dice by 1.7\% and 3.7\%, and disabling them together degrades Dice by 3.6\% and 7.3\% on these two sub-test sets, respectively.
This demonstrates that both short-term feature aggregation and long-term feature interaction are important for polyp video segmentation
Comparing the first row and the last two rows in Table.~\ref{table2}, removing either FA or MA results in performance degradation.
The reason is that the absence of FA makes it more difficult to reliably aggregate short-term features between adjacent frames with significant changes, and the absence of MA leads to the loss of polyp discriminative cues in historical frames, negatively affecting the reliability of shor-term and long-term interactions.

\section{Conclusion}
In this work, we proposed SALI, a novel efficient framework for polyp video segmentation.
SALI propose two novel modules, i.e., SAM and LIM, to address the stable shor-term aggregation of features between adjacent frames with large variations exits and reliable long-term temporal interaction over a plenty of low-quality frames, respectively.
The comprehensive and extensive comparisons with six state-of-the-art (SOTA) video segmentation method on SUN-SEG demonstrate that SALI outperforms the other SOTAs significantly.
In the future, we will investigate the label-friendly polyp video segmentation, such as requiring labels for only a few frames in a video or leveraging unlabeled videos to improve model performance.

\section*{Acknowledgements}
This work was supported in part by National Key \&  Program of China (Grant No. 2023YFC2414900), Key \& Program of Hubei Province of China (No.2023BC-B003), Fundamental Research Funds for the Central Universities (2021XXJS033), Wuhan United Imaging Healthcare Surgical Technology Co., Ltd.

%
%
%
%

\bibliographystyle{splncs04}
\bibliography{reference}

\begin{thebibliography}{10}
\providecommand{\url}[1]{\texttt{#1}}
\providecommand{\urlprefix}{URL }
\providecommand{\doi}[1]{https://doi.org/#1}

\bibitem{achanta2009frequency}
Achanta, R., Hemami, S., Estrada, F., Susstrunk, S.: Frequency-tuned salient region detection. In: 2009 IEEE conference on computer vision and pattern recognition. pp. 1597--1604. IEEE (2009)

\bibitem{chen2024mast}
Chen, G., Yang, J., Pu, X., Ji, G.P., Xiong, H., Pan, Y., Cui, H., Xia, Y.: Mast: Video polyp segmentation with a mixture-attention siamese transformer. arXiv preprint arXiv:2401.12439  (2024)

\bibitem{cheng2021rethinking}
Cheng, H.K., Tai, Y.W., Tang, C.K.: Rethinking space-time networks with improved memory coverage for efficient video object segmentation. Advances in Neural Information Processing Systems  \textbf{34},  11781--11794 (2021)

\bibitem{cheng2022implicit}
Cheng, X., Xiong, H., Fan, D.P., Zhong, Y., Harandi, M., Drummond, T., Ge, Z.: Implicit motion handling for video camouflaged object detection. In: Proceedings of the IEEE/CVF Conference on Computer Vision and Pattern Recognition. pp. 13864--13873 (2022)

\bibitem{dai2017deformable}
Dai, J., Qi, H., Xiong, Y., Li, Y., Zhang, G., Hu, H., Wei, Y.: Deformable convolutional networks. In: Proceedings of the IEEE international conference on computer vision. pp. 764--773 (2017)

\bibitem{deng2009imagenet}
Deng, J., Dong, W., Socher, R., Li, L.J., Li, K., Fei-Fei, L.: Imagenet: A large-scale hierarchical image database. In: 2009 IEEE conference on computer vision and pattern recognition. pp. 248--255. Ieee (2009)

\bibitem{dong2021polyp}
Dong, B., Wang, W., Fan, D.P., Li, J., Fu, H., Shao, L.: Polyp-pvt: Polyp segmentation with pyramid vision transformers. arXiv preprint arXiv:2108.06932  (2021)

\bibitem{fan2017structure}
Fan, D.P., Cheng, M.M., Liu, Y., Li, T., Borji, A.: Structure-measure: A new way to evaluate foreground maps. In: Proceedings of the IEEE international conference on computer vision. pp. 4548--4557 (2017)

\bibitem{fan2021cognitive}
Fan, D.P., Ji, G.P., Qin, X., Cheng, M.M.: Cognitive vision inspired object segmentation metric and loss function. Scientia Sinica Informationis  \textbf{6}(6) (2021)

\bibitem{fan2020pranet}
Fan, D.P., Ji, G.P., Zhou, T., Chen, G., Fu, H., Shen, J., Shao, L.: Pranet: Parallel reverse attention network for polyp segmentation. In: International conference on medical image computing and computer-assisted intervention. pp. 263--273. Springer (2020)

\bibitem{ji2021progressively}
Ji, G.P., Chou, Y.C., Fan, D.P., Chen, G., Fu, H., Jha, D., Shao, L.: Progressively normalized self-attention network for video polyp segmentation. In: International Conference on Medical Image Computing and Computer-Assisted Intervention. pp. 142--152. Springer (2021)

\bibitem{ji2022video}
Ji, G.P., Xiao, G., Chou, Y.C., Fan, D.P., Zhao, K., Chen, G., Van~Gool, L.: Video polyp segmentation: A deep learning perspective. Machine Intelligence Research  \textbf{19}(6),  531--549 (2022)

\bibitem{kingma2014adam}
Kingma, D.P., Ba, J.: Adam: A method for stochastic optimization. arXiv preprint arXiv:1412.6980  (2014)

\bibitem{lin2023shifting}
Lin, J., Dai, Q., Zhu, L., Fu, H., Wang, Q., Li, W., Rao, W., Huang, X., Wang, L.: Shifting more attention to breast lesion segmentation in ultrasound videos. In: International Conference on Medical Image Computing and Computer-Assisted Intervention. pp. 497--507. Springer (2023)

\bibitem{margolin2014evaluate}
Margolin, R., Zelnik-Manor, L., Tal, A.: How to evaluate foreground maps? In: Proceedings of the IEEE conference on computer vision and pattern recognition. pp. 248--255 (2014)

\bibitem{paszke2019pytorch}
Paszke, A., Gross, S., Massa, F., Lerer, A., Bradbury, J., Chanan, G., Killeen, T., Lin, Z., Gimelshein, N., Antiga, L., et~al.: Pytorch: An imperative style, high-performance deep learning library. Advances in neural information processing systems  \textbf{32} (2019)

\bibitem{pei2022hierarchical}
Pei, G., Shen, F., Yao, Y., Xie, G.S., Tang, Z., Tang, J.: Hierarchical feature alignment network for unsupervised video object segmentation. In: European Conference on Computer Vision. pp. 596--613. Springer (2022)

\bibitem{puyal2020endoscopic}
Puyal, J.G.B., Bhatia, K.K., Brandao, P., Ahmad, O.F., Toth, D., Kader, R., Lovat, L., Mountney, P., Stoyanov, D.: Endoscopic polyp segmentation using a hybrid 2d/3d cnn. In: Medical Image Computing and Computer Assisted Intervention--MICCAI 2020: 23rd International Conference, Lima, Peru, October 4--8, 2020, Proceedings, Part VI 23. pp. 295--305. Springer (2020)

\bibitem{teed2020raft}
Teed, Z., Deng, J.: Raft: Recurrent all-pairs field transforms for optical flow. In: Computer Vision--ECCV 2020: 16th European Conference, Glasgow, UK, August 23--28, 2020, Proceedings, Part II 16. pp. 402--419. Springer (2020)

\bibitem{vaswani2017attention}
Vaswani, A., Shazeer, N., Parmar, N., Uszkoreit, J., Jones, L., Gomez, A.N., Kaiser, {\L}., Polosukhin, I.: Attention is all you need. Advances in neural information processing systems  \textbf{30} (2017)

\bibitem{wang2022pvt}
Wang, W., Xie, E., Li, X., Fan, D.P., Song, K., Liang, D., Lu, T., Luo, P., Shao, L.: Pvt v2: Improved baselines with pyramid vision transformer. Computational Visual Media  \textbf{8}(3),  415--424 (2022)

\bibitem{wei2021shallow}
Wei, J., Hu, Y., Zhang, R., Li, Z., Zhou, S.K., Cui, S.: Shallow attention network for polyp segmentation. In: Medical Image Computing and Computer Assisted Intervention--MICCAI 2021: 24th International Conference, Strasbourg, France, September 27--October 1, 2021, Proceedings, Part I 24. pp. 699--708. Springer (2021)

\bibitem{wu2019cascaded}
Wu, Z., Su, L., Huang, Q.: Cascaded partial decoder for fast and accurate salient object detection. In: Proceedings of the IEEE/CVF conference on computer vision and pattern recognition. pp. 3907--3916 (2019)

\bibitem{yuan2023isomer}
Yuan, Y., Wang, Y., Wang, L., Zhao, X., Lu, H., Wang, Y., Su, W., Zhang, L.: Isomer: Isomerous transformer for zero-shot video object segmentation. In: Proceedings of the IEEE/CVF International Conference on Computer Vision. pp. 966--976 (2023)

\bibitem{zhang2021deep}
Zhang, K., Zhao, Z., Liu, D., Liu, Q., Liu, B.: Deep transport network for unsupervised video object segmentation. In: Proceedings of the IEEE/CVF International Conference on Computer Vision. pp. 8781--8790 (2021)

\bibitem{zhou2023cross}
Zhou, T., Zhou, Y., He, K., Gong, C., Yang, J., Fu, H., Shen, D.: Cross-level feature aggregation network for polyp segmentation. Pattern Recognition  \textbf{140},  109555 (2023)

\end{thebibliography}




\end{document}